\documentclass[10pt, a4paper]{article}

\usepackage[final]{lrec2026} 
\usepackage{amsmath,graphicx,hyperref, subcaption, stfloats, siunitx}
\usepackage{booktabs}
\usepackage{multirow}

\usepackage{xcolor}
\usepackage[dvipsnames]{xcolor}

\title{The Speech-LLM Takes It All: \\A Truly Fully End-to-End Spoken Dialog State Tracking Approach}

\name{Nizar El Ghazal\sthanks{The author performed the work while at an internship}, Antoine Caubrière, Valentin Vielzeuf} 

\address{Orange Research\\
         4 Rue du Clos Courtel, 35510 Cesson-Sévigné, France \\
         \texttt{firstname}.\texttt{lastname}@orange.com}

\abstract{
This paper presents a comparative study of context management strategies for end-to-end Spoken Dialog State Tracking using Speech-LLMs. We systematically evaluate traditional multimodal context (combining text history and spoken current turn), full spoken history, and compressed spoken history approaches. Our experiments on the SpokenWOZ corpus demonstrate that providing the full spoken conversation as input yields the highest performance among models of similar size, significantly surpassing prior methods. Furthermore, we show that attention-pooling-based compression of the spoken history offers a strong trade-off, maintaining competitive accuracy with reduced context size. Detailed analysis confirms that improvements stem from more effective context utilization.
 \\ \newline  }

\begin{document}

\maketitleabstract

\section{Introduction}
\label{sec:intro}

Dialog State Tracking (DST) is a vital component in task-oriented dialog (TOD) systems~\cite{suendermann2011slu,williamsDialogStateTracking2016}, enabling them to understand and maintain the context of a conversation over multiple turns. By accurately tracking user intents and relevant information, DST allows systems to reason over dialog states and effectively fulfill user requests.
However, in the context of spoken dialog, Spoken DST remains a relatively immature research area, with current system performance significantly lagging behind those achieved in written dialog scenarios~\cite{si2023spokenwoz}. 

One of the most common recent approaches 
is the cascade system. The process generally begins with an Automatic Speech Recognition (ASR) module that transcribes spoken language into text. This is often followed by an ASR correction module to improve the accuracy of the transcription, and then a written DST component, which is frequently based on models such as \textit{T5}~\cite{raffel2020exploring}. This pipeline approach leverages the strengths of existing text-based DST models. And it was notably popular in the DSTC-11 challenge~\cite{soltau-etal-2023-dstc}, where a variant was used by the winning system, OLISIA~\cite{jacqmin2023firstDSTC11}.

Despite its success, the cascade approach encounters inherent limitations. It is highly vulnerable to error propagation originating from the initial ASR stage. This can significantly affect the overall accuracy and reliability of the entire system~\cite{DSTErrors}.  This issue is even more pronounced in real-world scenarios, where ASR systems often struggle with proper nouns and domain-specific terminology, elements that are very frequent in DST slot values~\cite{Budzianowski2018MultiWOZA}.

End-to-end (E2E) systems have emerged as a promising alternative, as they may potentially mitigate the error propagation inherent in cascade systems.
In particular, \cite{druart2023one} demonstrated the effectiveness of E2E approaches, particularly in fully spoken contexts without access to ground-truth transcriptions, such as the SpokenWOZ~\cite{si2023spokenwoz} dataset. In these settings, E2E models have been shown to outperform traditional cascade systems.

Concurrently, speech-aware large language models (speechLLMs), which are also considered E2E systems, have gained increasing popularity in a variety of spoken language tasks, including ASR and response generation~\cite{ji2024wavchat,lu2025slide}. Recent work~\cite{sedlavcek2025approaching} applied speech-aware LLMs to the spoken DST task, achieving state-of-the-art performance in the SpokenWOZ dataset~\cite{si2023spokenwoz}.

A notable advantage of E2E systems is their remarkable flexibility when it comes to managing context, as they are capable of seamlessly integrating both written and spoken information within a single framework. For example, in recent studies such as \cite{druart2023one} and \cite{sedlavcek2025approaching}, both approaches utilize the spoken representation of the user's last turn to inform the system's response.

However, they differ significantly in how they handle the rest of the context: the former combines the spoken user turn with the written previous state, while the latter integrates it with the written representations of all previous turns, providing a large and more comprehensive context. This difference in approach raises an important and intriguing question. 

What would happen if we relied solely on spoken context? Specifically, what would be the effects of feeding the system the speech representations for the entire conversation, or alternatively, condensing these spoken representations using an intermediate module before processing?

In this paper, we explore these possibilities for context management when using a Speech-LLM model. 
Our contributions are three-fold.

\begin{itemize}
    \item We validate the use of Speech-LLMs as an accurate approach for spoken DST.
    \item We propose two context management approaches reaching the SOTA.
    \item Our best performing approach demonstrates a simple yet effective method: feeding the entire spoken conversation to the model without additional compression or modality mixing.
\end{itemize}

\section{Methodology}
\label{sec:methods}
In task-oriented dialog (TOD) systems, the role of Spoken Dialog State Tracking (DST) is to condense the user's intent and relevant information into a structured, machine-readable format. 
More formally, given as input a sequence of spoken dialog turns $U_1$, $A_2$, ..., $A_{t-1}$, $U_{t-1}$, our goal is to predict a set of $k$ relevant domains ($domain_1$, $domain_2$, ..., $domain_k$) and $n$ slot-value pairs ($slot_1=value_1$, $slot_2=value_2$, ..., $slot_n=value_n$), which are then represented as a JSON structure.

Figure~\ref{fig:system_diagram} illustrates our proposed systems, composed of three main components: a speech encoder, a connector, and a Large Language Model (LLM). In order to reduce the context length, we optionally add a "compression module" between the connector and LLM. The speech encoder processes the entire dialog history and computes dense representations for each turn. These representations are then down-sampled, using x6 stride, and passed to the connector module, which maps the speech features into the LLM's input space. They may be passed through the compression module for the approaches that need it. Finally, the LLM generates the dialog state in an auto-regressive manner.

\subsection{Context Management}
\label{subsec:context_management}
\begin{figure*}[ht]
    \centering
    \includegraphics[width=\linewidth]{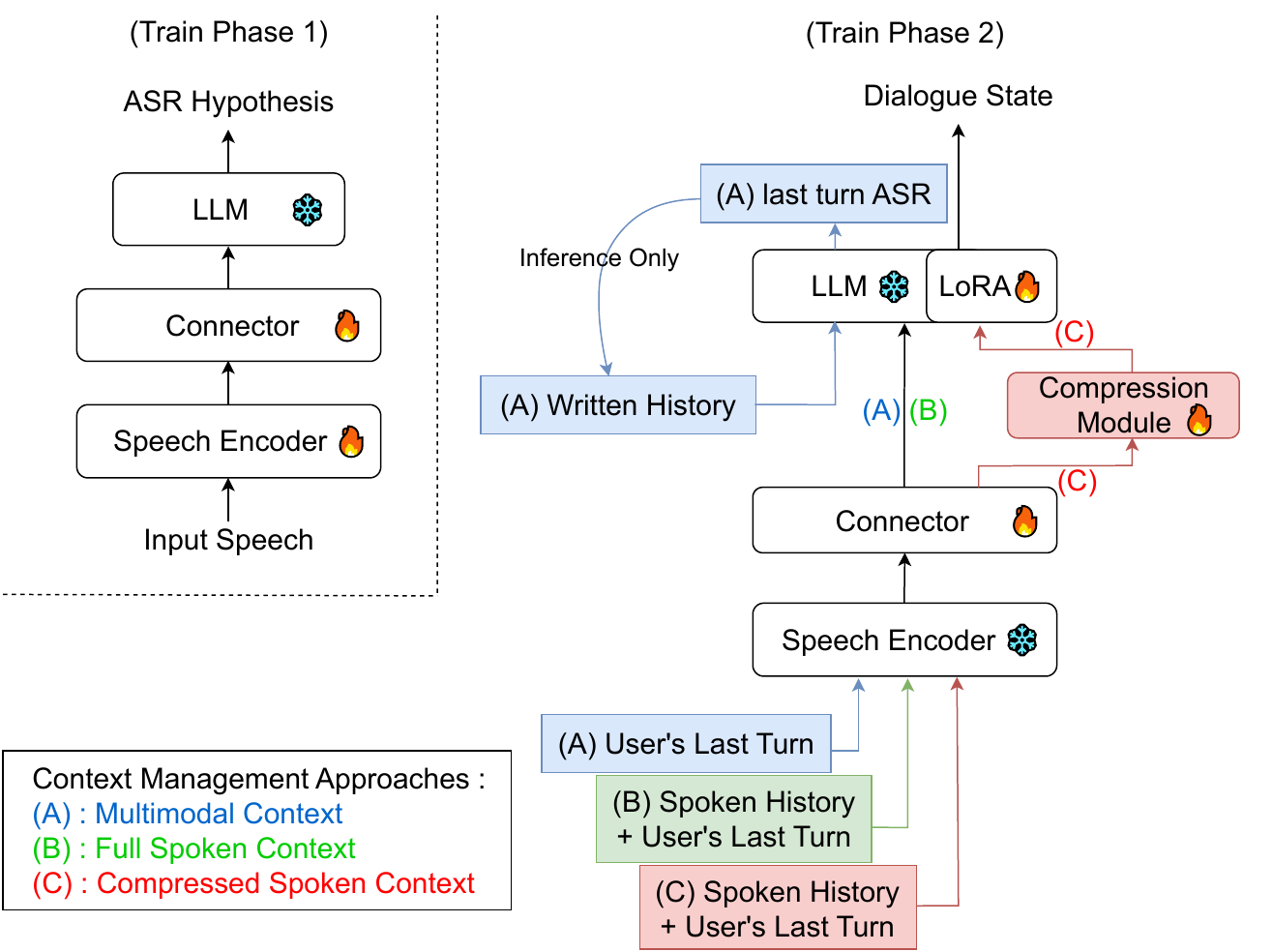}
    \caption{An overview of our system. to the left, the ASR pretraining stage. To the right finetuning for dialog state tracking}
    \label{fig:system_diagram}
\end{figure*}

As represented in Figure~\ref{fig:system_diagram}, we explore several strategies for handling the dialog context.

\paragraph*{Multimodal Context}
Following \cite{sedlavcek2025approaching}, we provide as input the spoken user utterance $U^{\text{spoken}}_n$ and the written dialog history together. The model then predicts the transcription of the user's utterance $\textcolor{teal}{U^{text}_n}$, the active domains $\textcolor{RoyalPurple}{D_n}$ and the dialog state \textcolor{orange}{$S_n$}.
The LLM is trained on the prompt:

\noindent
\fbox{
\begin{minipage}{0.95\linewidth}
\textit{\textcolor{Rhodamine}{$h_n$}
\{ "history": \textcolor{brown}{$Context_n$}\,,
    "user\_last\_turn": \textcolor{teal}{$U^{text}_n$}\,,
    "domains": \textcolor{RoyalPurple}{$D_n$}\,,
    "predicted\_state": \textcolor{orange}{$S_n$}\ \} }
\end{minipage}
}

where we have:
\begin{equation*}
\begin{aligned}
\mathcolor{brown}{Context_n} &=
\texttt{USER: } U_1 \; ;\\
&\quad \texttt{AGENT: } A_2 \; ;\\
&\quad \dots \; ;\\
&\quad \texttt{AGENT: } A_{n-1}
\end{aligned}
\\
\end{equation*}
\begin{equation*}
\mathcolor{Rhodamine}{h_n} = Connector\big(Encoder(U_n)\big)
\end{equation*}
    
In practice, the speech representation $h_n$ is concatenated with embeddings that represent the prompt’s text, yielding a multimodal input sequence. 
During inference, the model autoregressively completes the prompt starting 
from the field \texttt{"user\_last\_turn"}. The generated ASR hypothesis 
$\textcolor{teal}{U^{text}_n}$ is then fed back to construct the textual context 
\textcolor{brown}{$Context_{n+1}$} for subsequent turns.

\paragraph*{Full Spoken Context}
With this context-management strategy, $\text{Context}_n$, corresponding to the full spoken conversation, is provided to the model. The model predicts the active domain $\textcolor{RoyalPurple}{D_n}$ and the dialog state \textcolor{orange}{$S_n$}.
The prompt employed for this strategy is:\\

\noindent
\fbox{
\begin{minipage}{0.95\linewidth}
\textit{\textcolor{Rhodamine}{Speech\_Emb}
    \{"domains": \textcolor{RoyalPurple}{$D_n$}\,,"predicted\_state": \textcolor{orange}{$S_n$}\ \}}
\end{minipage}
}

where :
\begin{equation*}  
\begin{aligned}
\textcolor{brown}{Context_n} &= (
    U^{\text{spoken}}_1,
    A^{\text{spoken}}_2,
    \dots,
    U^{\text{spoken}}_n
    ) \\
h_{2i+1} &= Connector\big(Encoder(U_{2i+1})\big) \\
h_{2i} &= Connector\big(Encoder(A_{2i})\big) \\
\textcolor{Rhodamine}{Speech\_Emb} &= (h_1 || h_2 || \dots || h_n)
\end{aligned}
\end{equation*}
As in the multimodal context setting, the sequence of speech embeddings \textcolor{Rhodamine}{$Speech\_Emb$} is pre-pended to the embeddings of the textual part of the prompt before being fed to the LLM. During inference, the model receives the speech embeddings as input and auto-regressively generates the remaining fields of the prompt.

\paragraph*{Compressed Spoken Context}
The only difference with full spoken context is how \textcolor{Rhodamine}{$Speech\_Emb$} is obtained.
Instead of using the entire sequences $h_i$, we introduce a set of 
$N_{\text{queries}}$ trainable query vectors $Q$ and compute $z_i$ through query-based pooling using a TransformerDecoder architecture:
\begin{align*}
z_i &= \text{TransformerDecoder}(Q, h_i) \\
\textcolor{Rhodamine}{Speech\_Emb} &= (z_1 || z_2 || \dots || h_n)
\end{align*}
In this formulation, the decoder treats $Q$ as the target sequence 
and $z_i$ as the memory. Each decoder layer first applies 
\emph{self-attention} over the query tokens, allowing them to 
interact and share information. It then applies 
\emph{cross-attention}, where the queries attend to the speech 
sequence $z_i$, extracting the most relevant aspects from it. 
The final output is a set of $N_{\text{queries}}$ vectors 
that serve as a compressed representation of the turn. 
These vectors are concatenated and used in downstream 
dialog modeling.
\subsection{Training}
\label{subsec:trainig}
We train our models in two stages, as described in Figure~\ref{fig:system_diagram}. The first stage is ASR pre-training, where we freeze the LLM and train the speech encoder and connector to produce speech representations that align with the LLM's input space. Specifically, we task the LLM with generating the transcription from the speech embeddings, propagating the LLM gradients back to the encoder and connector. This approach allows us to leverage the large-scale ASR datasets that are publicly available, resulting in robust alignment between the speech and text modalities.

The second stage is DST fine-tuning. In this phase, we freeze the speech encoder and train the connector, the optional compression module, and a small LoRA module for the LLM. The objective is to produce a JSON string in the format described in~\ref{subsec:context_management}. Training is performed by minimizing the cross-entropy loss between the generated output and the ground-truth dialog state annotations.

\section{Results}
\label{sec:results}

\subsection{Datasets}
\label{subsec:datasets}
For the ASR pre-training stage, we train our model on a combination of the Loquacious Medium dataset (2,500 hours)~\cite{parcollet2025loquaciousset25000hours}, the Fisher corpus (1,960 hours)~\cite{cieri2004fisher}, and the train split from SpokenWOZ dataset (200 hours)~\cite{si2023spokenwoz}. Although SpokenWOZ does not provide ground-truth transcripts, we include it in the ASR pre-training phase because the speech encoder is frozen during DST fine-tuning, and we want the encoder to be exposed to the characteristics of SpokenWOZ data. To address the lack of transcripts on SpokenWOZ, we use Whisper-large-v3\footnote{https://huggingface.co/openai/whisper-large-v3}~\cite{radford2023robust} to generate automatic transcriptions for SpokenWOZ audio. These generated transcripts are also used later for the multimodal context method in the DST stage.

For DST fine-tuning, we primarily use the SpokenWOZ dataset for both training and evaluation. As in \cite{druart2023one,sedlavcek2025approaching} we remove the nine corrupted dialogs from the SpokenWOZ test set\footnote{https://github.com/AlibabaResearch/DAMO-ConvAI/issues/87}, and report the Joint Goal Accuracy (JGA)~\cite{zhong-etal-2018-global} on both the dev and test sets.
\begin{table*}[h]
\centering
\begin{tabular}{lc}
\hline
Model & SpokenWOZ test \\ \hline
SPACE+WavLMalign \cite{si2023spokenwoz} & 25.65\% \\
E2E (Whisper+T5) \cite{druart2023one} & 24.10\% \\
UBAR + GenWOZ \cite{gulzar2025leveraging} & 25.90\% \\
WavLM + conn. + OLMo-1B \cite{sedlavcek2025approaching} & 34.66\% \\
Compressed Spoken Context (Ours) & 36.49\% \\
Full Spoken Context (Ours) & \textbf{39.32\%} \\ \hline
WavLM + conn. + Gemma2-9B-Instruct \cite{sedlavcek2025approaching} & 42.17\% \\
Compressed Spoken Context + Gemma2-9B-Instruct (Ours) & 43.16\% \\
Full Spoken Context + Gemma2-9B-Instruct (Ours) & \textbf{45.52\%} \\
\end{tabular}
\caption{Comparison of our approaches with prior works in two different setups: using open data models (upper part) and using potentially data-contaminated models such as Gemma2-9B-Instruct (bottom).}
\label{tab:sota}

\end{table*}
\subsection{Implementation details}
\label{subsec:implementation}
For our component selection, we use W2v-BERT~\footnote{https://huggingface.co/facebook/w2v-bert-2.0}~\cite{communication2023seamlessmultilingualexpressivestreaming} as the speech encoder. The connector module is implemented as a single-layer Transformer encoder with a hidden dimension of 1024 and 16 attention heads. Similarly, we employ a one-layer Transformer Decoder with a hidden dimension of 1024, 16 heads, and a trainable number of queries ($N_{queries}$) as the compression module. This module is also used for attention pooling by setting $N_{queries} = 1$.

For the language model, we use OLMo 2 1B\footnote{https://huggingface.co/allenai/OLMo-2-0425-1B-Instruct}~\cite{olmo20252olmo2furious}. We apply a LoRA adapter with a rank of 16 and an alpha value of 1, as determined by grid search. 
Note that for the specific case of scaling experiments, we follow the protocol of \cite{sedlavcek2025approaching} and we therefore train new variants of our systems using W2V-BERT-2 as the speech encoder and Gemma2-9B-Instruct\footnote{https://huggingface.co/google/gemma-2-9b-it}as the LLM, while keeping the same connector. The training ASR datasets remained unchanged, but we used the augmented Speech Aware MultiWOZ~\cite{soltau-etal-2023-dstc} with SpokenWOZ for model training, followed by a final finetuning step on SpokenWOZ alone.

During inference, we employ beam search with 5 beams, which was also selected based on grid search results.
During ASR pre-training, we use a virtual batch size of 256, a learning rate of $1 \times 10^{-4}$, and 5,000 warm-up steps. Training proceeds until the word error rate (WER) on the combined validation sets of all datasets ceases to improve.
For DST fine-tuning, we maintain the same virtual batch size of 256, use a learning rate of $2 \times 10^{-4}$, and 500 warm-up steps. The model is trained until the JGA on the validation set no longer improves.
All our experiments\footnote{The source code will be released after acceptance} were performed using SpeechBrain toolkit\footnote{https://github.com/speechbrain/speechbrain}~\cite{speechbrain_v1}

\subsection{Best Model Analysis}
\label{subsec:best_analysis}
For fair comparison with prior work, the reported JGA for our model in Table \ref{tab:sota} uses post-processing, which includes (i) canonicalizing time expressions to 24-hour format and (ii) case-insensitive fuzzy matching for open/proper-noun slots with a Levenshtein~ratio~$\geq$~0.90, applied symmetrically to predictions and references.

Table~\ref{tab:sota} presents a comparison between published results on the SpokenWOZ test set and our two best systems: the compressed context method using 10 queries and the full spoken context method. For our systems, the post-processing yields a 3 points JGA increase, which is comparable to the post-processing reported in \cite{sedlavcek2025approaching}. 

Our approach substantially outperforms other systems of comparable size. Note that we distinguish two different setups. (a) For the upper part of Table~\ref{tab:sota}, we focus on approaches using LLMs with known and identified training data, guaranteeing the absence of test data contamination. And (b) for the bottom part, we report performance for sake of comparison with \cite{sedlavcek2025approaching}, which is using Gemma2-9B LLM. As this model training set is not known and according to \cite{sedlavcek2025approaching}, we can't be sure that there is no overlap with the test data of SpokenWOZ. Moreover, the number of parameters is larger than for upper lines methods, allowing to draw comparison between approaches for larger scale models. Note that the training data of these larger models in the specific DST step are following the protocol of \cite{sedlavcek2025approaching}, provided in the implementation details subsection \ref{subsec:implementation}.


\begin{figure*}[hb]
    \centering    \includegraphics[width=\linewidth]{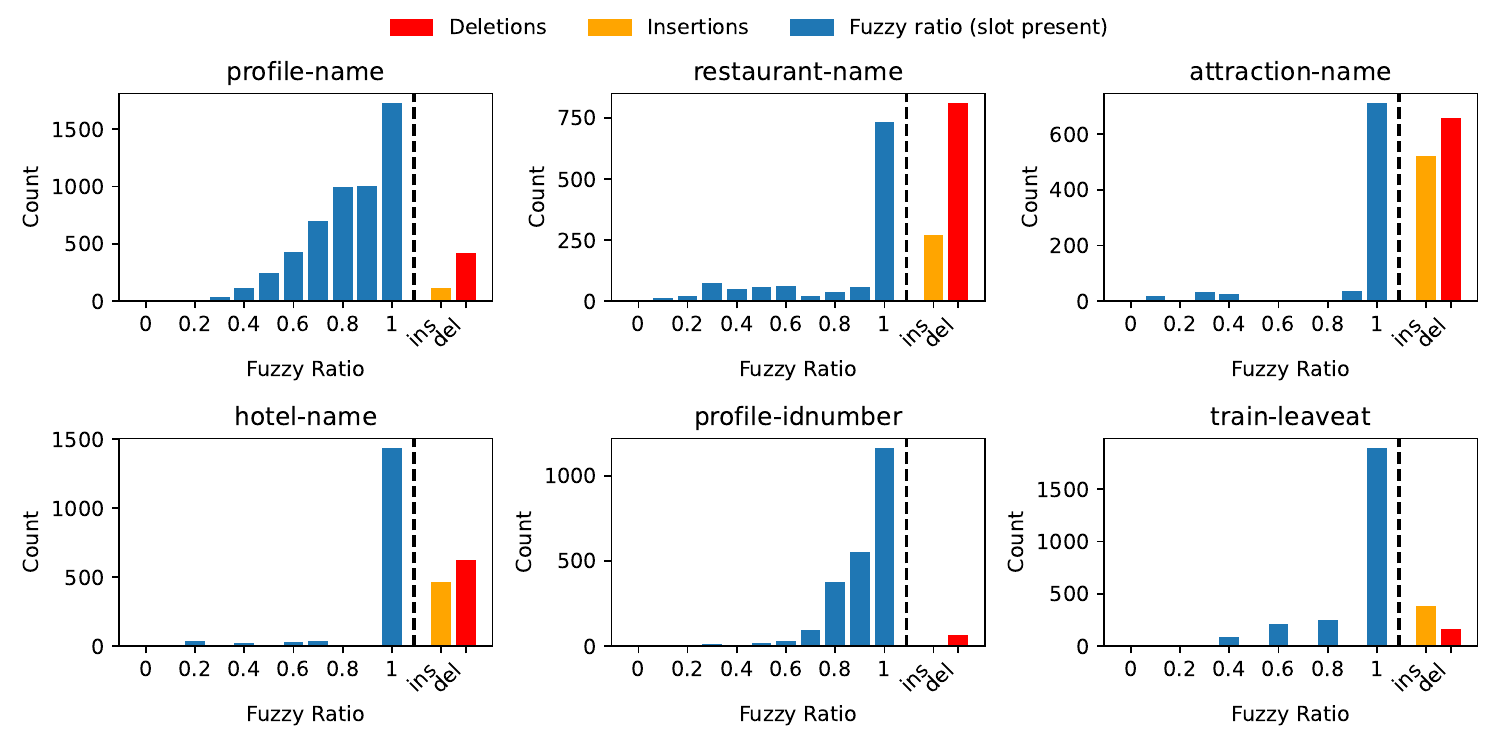}
    \caption{Distribution of Levenshtein (fuzzy) ratios for the six most error-prone slots, with counts of insertions (orange) and deletions (red). High fuzzy ratios indicate near-correct predictions.}
    \label{fig:best_errors}
\end{figure*}
To further analyze our best model, we selected the six slots with the highest error counts. In Figure~\ref{fig:best_errors}, blue bars represent the Levenshtein (fuzzy) ratio for slot values present in both prediction and reference, while orange and red bars indicate the counts of insertions and deletions, respectively. Most predictions achieve high fuzzy ratios (above 0.8), suggesting that when the model predicts a slot present in the reference, it usually gets the value nearly correct.
Interestingly, for \texttt{restaurant-name}, \texttt{attraction-name}, and \texttt{hotel-name}, the number of substitutions (fuzzy ratio $<$ 1) is very low, with most errors arising from insertions and deletions. This indicates that the model is generally able to correctly predict these proper nouns when it attempts them. In contrast, profile-related slots (e.g., \texttt{profile-name}, \texttt{profile-idnumber}) remain highly challenging due to their variable content and frequent spelling across multiple turns.
Finally, although the error rate for \texttt{train-leaveat} is relatively low compared to its total occurrences, its high frequency means it still contributes substantially to the overall error count.

\subsection{Context Management Methods Comparison}
\label{subsec:method_comparison}
All subsequent analyses use JGA without post-processing.
Table \ref{tab:method_comparison} shows the JGA score on SpokenWOZ dev and test splits for each method. Overall, both the full spoken context and the 10-queries-per-turn methods outperformed the baseline. In particular, the full spoken context approach achieved a significantly higher JGA, demonstrating the effectiveness of leveraging the entire spoken conversation as input. The competitive performance of the 10-queries method further suggests that a substantial portion of the speech representations is redundant, and that it is possible to reduce the input size without a significant loss in performance, provided that a sufficient number of queries is used. We next provide a fine-grained comparison based on slot group and dialog turn analyses.
\begin{table}[ht] 
\centering 
\begin{tabular}{lcc}
\hline
 Context                              & Dev             & Test            \\ \hline
Multimodal (baseline)  & 31.85\%              & 32.06\%              \\
Full Spoken             & \textbf{36.89\%}     & \textbf{36.29\%}     \\
Compressed Spoken      & \multicolumn{1}{l}{} & \multicolumn{1}{l}{} \\
\multicolumn{1}{r}{1 query}    & 31.03\%              & 30.99\%              \\
\multicolumn{1}{r}{10 queries} & 34.26\%              & 33.51\%              \\ \hline
\end{tabular}
\caption{JGA Evaluation of different context management approaches on SpokenWOZ.} \label{tab:method_comparison} 
\end{table}

\begin{figure}[ht]
    \centering
    \begin{subfigure}{\linewidth}
        \centering
        \includegraphics[width=\linewidth]{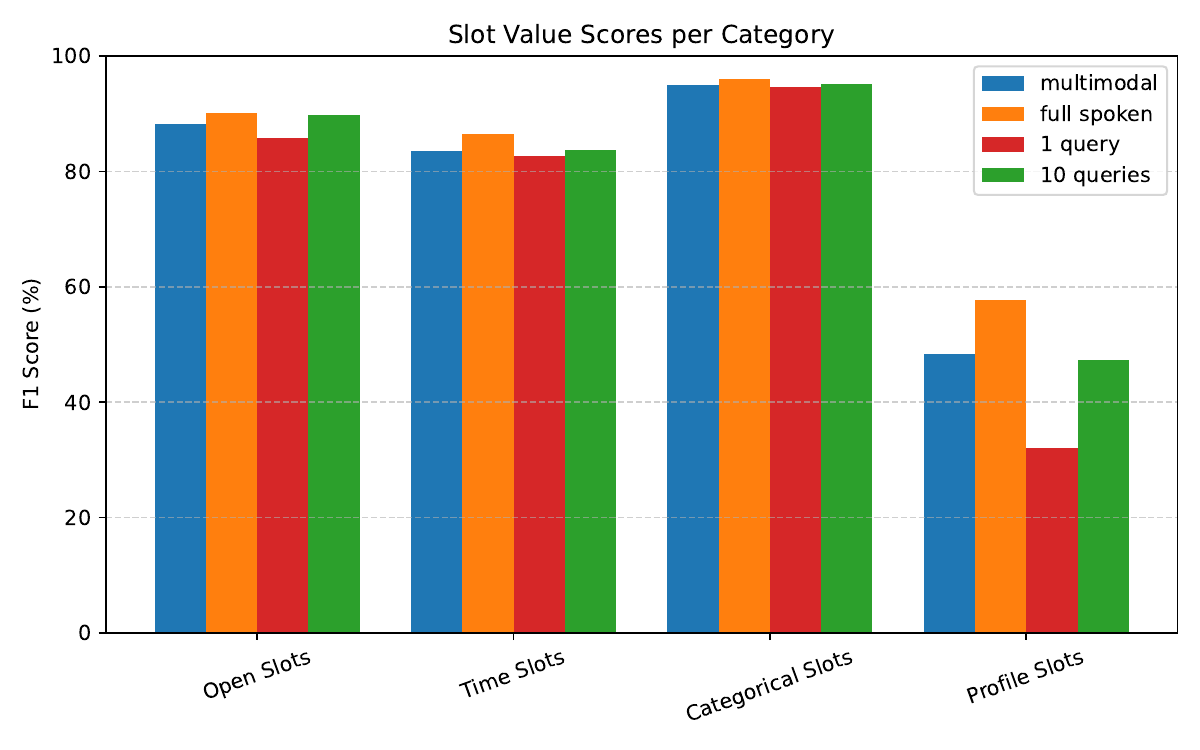}
        \caption{Slot type analysis}
        \label{fig:slot_type_analysis}
    \end{subfigure}
    \hfill
    \begin{subfigure}{\linewidth}
        \centering
        \includegraphics[width=\linewidth]{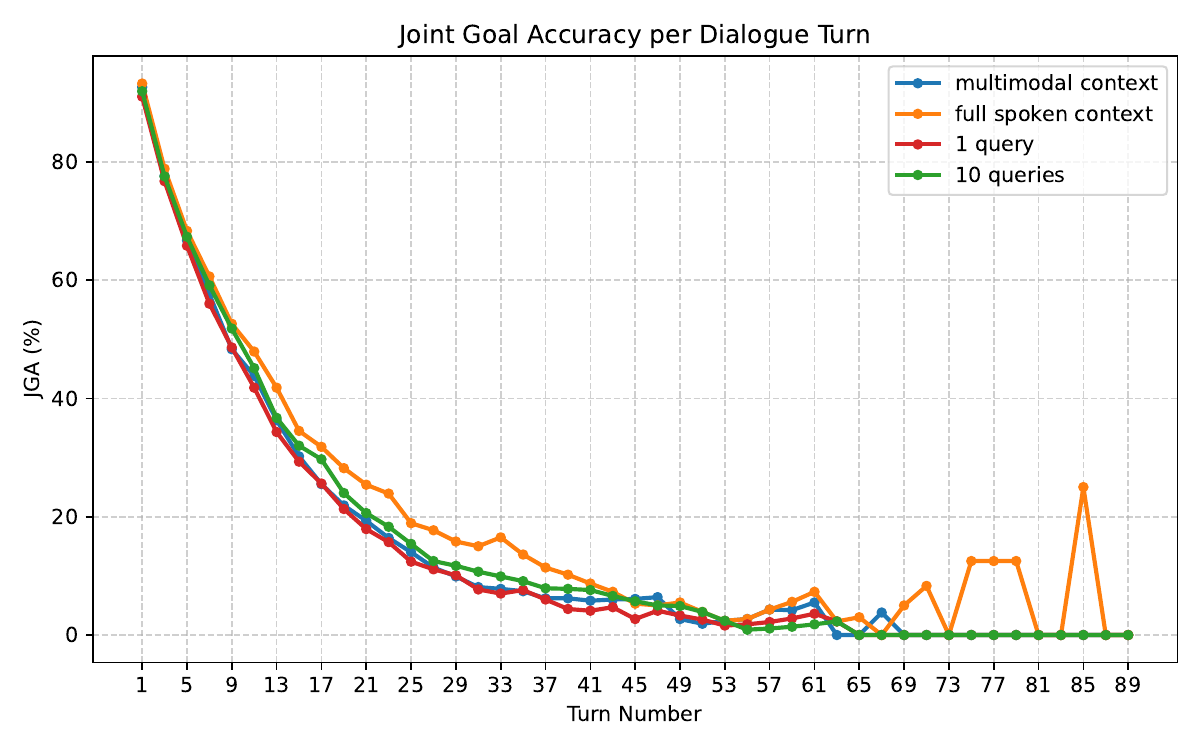}
        \caption{JGA per turn}
        \label{fig:jga_per_turn_analysis}
    \end{subfigure}
    \caption{(a) Slot value F1 score analysis per category. (b) JGA score analysis per dialog turn.}
    \label{fig:slot_and_turn_analysis}
\end{figure}

\paragraph*{Slot Group Analysis}
We categorize slots into four groups: categorical, time, open, and profile. Categorical slots have a fixed set of values (e.g., yes/no, area, price range). Time slots correspond to temporal expressions (e.g., departure time). Open slots can take a wide range of values such as place names, while profile slots, which are treated separately for finer analysis, contain personal information (e.g., names, IDs, emails) and are often spelled out across multiple turns.
Figure~\ref{fig:slot_type_analysis} shows the average F1 score by slot type. All models perform well on categorical slots, with full spoken context slightly ahead. Performance drops for time and open slots, where full spoken context and 10-query compression clearly outperform the others. Profile slots are the hardest: full spoken context again leads, while the 1-query model performs worst, indicating that compressing each turn to a single embedding discards too much information.

\paragraph*{Dialog Turn Analysis}
Figure~\ref{fig:jga_per_turn_analysis} displays the evolution of Joint Goal Accuracy (JGA) across dialog turns. All models perform well in the early turns (1–5), but accuracy declines quickly in the mid turns (5–30) and approaches zero by turn 40. This drop can be attributed to the increasing length and complexity of dialog states, combined with the strictness of the JGA metric, as well as the limited capacity of the relatively small LLM used in our experiments.
The full spoken context method consistently outperforms the others, particularly during the mid turns. In the very late turns, it shows occasional performance peaks, though these are difficult to interpret given the small sample size. The 10-query attention pooling method remains competitive, but still underperforms compared to full spoken context in the late turns, even though it benefits from a much smaller context size.

\subsection{Additional Experiences}
\label{subsec:ablations}

\paragraph*{Scaling model and data}
\begin{table*}[ht]
\centering
\begin{tabular}{lll|cc}
\hline
\multicolumn{1}{l|}{\multirow{2}{*}{}}                                                & \multicolumn{1}{l|}{\multirow{2}{*}{LLM}}       & \multirow{2}{*}{Train Data} & \multicolumn{2}{c}{SpokenWOZ Test}                           \\ \cline{4-5} 
\multicolumn{1}{l|}{}                                                                 & \multicolumn{1}{l|}{}                           &                             & \multicolumn{1}{c|}{JGA}                      & JGA+PP  \\ \hline
\multicolumn{1}{l|}{\multirow{5}{*}{Full Spoken Context}}                             & \multicolumn{1}{l|}{\multirow{2}{*}{OLMo2-1B}}  & SpokenWOZ                   & \multicolumn{1}{c|}{36.29\%}                  & 39.32\% \\
\multicolumn{1}{l|}{}                                                                 & \multicolumn{1}{l|}{}                           & Combined + SW ft            & \multicolumn{1}{c|}{36.23\%}                      & 39.14\%        \\ \cline{2-5} 
\multicolumn{1}{l|}{}                                                                 & \multicolumn{1}{l|}{\multirow{3}{*}{Gemma2-9B}} & SpokenWOZ                   & \multicolumn{1}{c|}{39.62\%}                  & 43.12\% \\
\multicolumn{1}{l|}{}                                                                 & \multicolumn{1}{l|}{}                           & Combined                    & \multicolumn{1}{c|}{40.06\%}                  & 43.74\% \\
\multicolumn{1}{l|}{}                                                                 & \multicolumn{1}{l|}{}                           & Combined + SW ft            & \multicolumn{1}{c|}{\textbf{41.99\%}}  & \textbf{45.52\%}        \\ \hline
\multicolumn{1}{l|}{\multirow{5}{*}{10 queries Compressed Spoken Context}} & \multicolumn{1}{l|}{\multirow{2}{*}{OLMo2-1B}}  & SpokenWOZ                   & \multicolumn{1}{c|}{33.51\%}                  & 36.49\% \\
\multicolumn{1}{l|}{}                                                                 & \multicolumn{1}{l|}{}                           & Combined + SW ft            & \multicolumn{1}{c|}{35.16\%}                  & 38.59\% \\ \cline{2-5} 
\multicolumn{1}{l|}{}                                                                 & \multicolumn{1}{l|}{\multirow{3}{*}{Gemma2-9B}} & SpokenWOZ                   & \multicolumn{1}{c|}{35.94\%}                  & 39.13\% \\
\multicolumn{1}{l|}{}                                                                 & \multicolumn{1}{l|}{}                           & Combined                    & \multicolumn{1}{c|}{37.01\%}                  & 40.15\% \\
\multicolumn{1}{l|}{}                                                                 & \multicolumn{1}{l|}{}                           & Combined + SW ft            & \multicolumn{1}{c|}{39.53\%} & 43.16\%       \\ \hline
\multicolumn{3}{l|}{WavLM   + conn. + Gemma2-9B-Instruct}                                                                                                            & \multicolumn{1}{c|}{38.76\%}                  & 42.17\% \\ \hline
\end{tabular}
\caption{JGA performance of our scaled models, JGA+PP uses the post-processing.}
\label{tab:scaling}
\end{table*}
Table~\ref{tab:scaling} shows the scaling results. We note that for the full spoken context on Gemma2-9B, we decided to introduce context truncation, limiting the total speech-embedding tokens passed to the LLM to 1500 at most, mainly in order to run the experiments with the available resources.
We observe that scaling the amount of train data is beneficial in all our setups (except the full spoken context with OLMo2-1B). However, the Full Spoken Context method benefits less compared to the Compressed Spoken Context. One possible reason for this is that the Transformer-Decoder introduced with this method is only initialized at the DST-finetuning stage, and SpokenWOZ alone is not able to fully train it to compress the context efficiently, which means that more data are beneficial for it, the same way as with the connector, which greatly benefits from more data during the ASR-pretraining.
We also observe that the bigger LLM gets more benefit from scaling the training data, which is expected since a bigger model needs more training data.
Finally, our Speech-LLM model with compressed spoken context is slightly ahead compared to the current SOTA while using a much smaller context size, which means less resources (especially since the attention scales quadratically with the context size). Finally, our Speech-LLM setup with the (truncated) full spoken context significantly outperforms the Gemma2-9B variant from~\cite{sedlavcek2025approaching}, establishing the new SOTA for Spoken DST on SpokenWOZ.

\paragraph*{Impact of the number of queries}
\begin{figure}[hb]
    \centering
    \includegraphics[width=\linewidth]{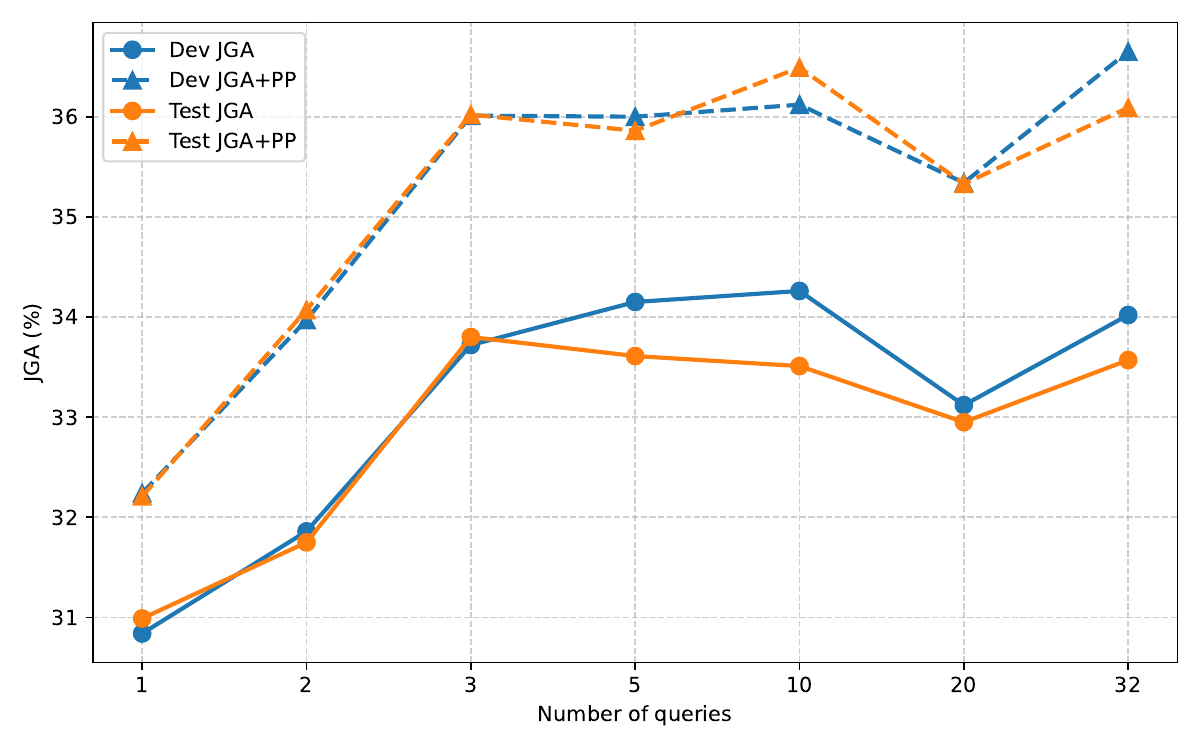}
    \caption{Impact of the number of queries on the JGA scores of the compressed spoken context systems.}
    \label{fig:jga_per_nqueries}
\end{figure}

Figure \ref{fig:jga_per_nqueries} shows the impact of the number of queries on performance. We observe that the JGA scores increase significantly when increasing the number of queries from 1 to 3, but remain pretty similar for higher number of queries. we also varied the number of layers and found that increasing to three layers led to a 2\% absolute drop in JGA. We attribute this to the limited amount of DST finetuning data, as the compression module is only initialized at this stage. 

\paragraph*{Other ablations} To better understand the contributions of individual components and design choices in our system, we conducted a series of ablation studies and supplementary experiments. Specifically, we investigated the impact of ASR pretraining data, the connector, the compression module, and DST data preprocessing. For ASR pretraining, we compared using the LibriSpeech dataset~\cite{panayotov2015librispeech} alone versus the mixed dataset described in Section~\ref{subsec:datasets}. 

In baseline experiments with the multimodal method, we observed that when the encoder is unfrozen during DST finetuning, the choice of ASR pretraining data has little impact. However, when freezing the encoder (which is a more practical setup for the Full/Compressed Spoken Context methods), we found that relying solely on LibriSpeech resulted in up to a 3-point drop in JGA compared to using the mixed dataset. 

During ASR pretraining, we also experimented with different numbers of layers (1, 2, and 4) in the encoder. We found that a single layer provided the fastest convergence and the best performance. 

Finally, for the multimodal context method, we normalized Whisper transcripts using NeMo Inverse Text Normalization (ITN)~\cite{zhang2021nemo}, along with additional processing for time expressions. This preprocessing yielded a 1\% absolute gain in JGA.

\section{Conclusion}
\label{sec:conclusion}
In this paper, we have proposed a fully E2E approach to Spoken Dialog State Tracking, drawing inspiration from Speech-LLMs.
In contrast to traditional multimodal context approaches, we show that it is possible to use the entire spoken conversation as input (until the current turn) and achieve state-of-the-art results.
We also performed a fine-grained analysis to illustrate the causes of improvements brought by using a full spoken context: less error propagation through the dialog and better performance on the most challenging slots. 

At the same time, using the full spoken history entails practical trade-offs. Most notably, the increased memory usage and latency motivate the design of more efficient context representations.
In future work, a more sophisticated and compact handling of the spoken context may be explored. Moreover, scaling the used model would be a promising extension. We further plan to investigate streaming inference and integration within production pipelines. We also are interested in enlarging the scope of the evaluation beyond SpokenWOZ, as additional human-spoken DST benchmarks may become available. 
Finally, we aim to release code and trained models to facilitate reproducibility.

\section{Limitations} 
While our experiments show that providing the full spoken history is beneficial for spoken DST, some limitations need to be mentioned. 
First, the memory and compute footprint of the full-spoken setting grows with dialog length. Indeed, each additional spoken turn expands the input sequence to process. It then increases attention cost (quadratically) and thus end-to-end latency. This problem is especially noticeable on long dialogs or when having production-level constraints on hardware (GPU memory can become a bottleneck even with relatively small context windows). We can note that our proposed compression variant helps tackle this issue, yet it may imply the loss of some fine-grained cues, such as named entities, numbers, or accurate time slots). 

Then our models are effective enough for an offline research environment. Yet, they remain far from a delivery-level. Real-world TOD systems often require to enable streaming (e.g. partial hypotheses, incremental state updates) with small latency. The scope of the paper was not including the end-to-end latency and the failure modes under streaming constraints, which are all prerequisites for reliable deployment and may be included in future work.

Finally, our empirical analysis is focused on SpokenWOZ. To the best of our knowledge, it is the main benchmark built from human conversational speech for DST. We may extend the study to other benchmarks such as Spoken MultiWOZ, but it would require care in interpretation. This kind of corpora is mainly composed of synthesized voice for training and validation. Human speech is only in evaluation splits. So a large part of the dataset contains acoustic  characteristics that are very different from fully human-spoken data. Thus, performance measured on such datasets may reflect robustness to TTS artifacts as much as to real conversational variability. 

Future works may help to address these three limitations. We may explore more memory-efficient context representations (such as hierarchical or selective retention), dig on streaming compatible inference, and enlarge the evaluation scope to synthesized corpora with great care for acoustic quality.

\section{Bibliography}
\bibliographystyle{lrec2026-natbib}
\bibliography{lrec2026-example}

\end{document}